\documentclass[runningheads]{llncs}
\usepackage{graphicx}
\usepackage{array,makecell}
\usepackage{tabularx}
\usepackage{multirow}
\usepackage{tikz}
\usepackage{comment}
\usepackage{amsmath,amssymb} 
\usepackage{color}
\usepackage{CJKutf8} 
\usepackage{CJK}
\usepackage{mathtools}
\usepackage{wrapfig} 
\usepackage{graphicx} 
\newcolumntype{P}[1]{>{\centering\arraybackslash}p{#1}}
\usepackage{float}

\usepackage{booktabs}  
\usepackage{threeparttable} 
\usepackage{enumitem}

\usepackage[accsupp]{axessibility}  

\begin{document}
\pagestyle{headings}
\mainmatter
\def\ECCVSubNumber{4996}  

\title{Adversarial Contrastive Learning via Asymmetric InfoNCE}

\newcommand{\jj}[1]{\textcolor{red}{\small{\bf [JJ: #1 ]}}}
\newcommand{\zhan}[1]{\textcolor{blue}{\small{\bf [zhan: #1]}}}

\titlerunning{Adversarial Contrastive Learning via Asymmetric InfoNCE}
%
\newcommand*\samethanks[1][\value{footnote}]{\footnotemark[#1]}
\author{
Qiying Yu\inst{1,2}\thanks{Corresponding authors}
\and
Jieming Lou\inst{2}
\and
Xianyuan Zhan\inst{1}
\and
Qizhang Li\inst{2}
\and
Wangmeng Zuo\inst{2}
\and
Yang Liu\inst{1,3}
\and
Jingjing Liu\inst{1}\samethanks[1]
}
\authorrunning{Q. Yu et al.}
%
\institute{
Institute for AI Industry Research, Tsinghua University, China
\and
School of Computer Science and Technology, Harbin Institute of Technology, China
\and
Department of Computer Science and Technology, Tsinghua University, China
\email{yuqy22@mails.tsinghua.edu.cn, jjliu@air.tsinghua.edu.cn}
}
\maketitle

\begin{abstract}
Contrastive learning (CL) has recently been applied to adversarial learning tasks. Such practice considers adversarial samples as additional positive views of an instance, and by maximizing their agreements with each other, yields better adversarial robustness. However, this mechanism can be potentially flawed, since adversarial perturbations may cause instance-level \textit{identity confusion}, which can impede CL performance by pulling together different instances with separate identities. To address this issue, we propose to treat adversarial samples unequally when contrasted, with an asymmetric InfoNCE objective (\textit{A-InfoNCE}) that allows discriminating considerations of adversarial samples. Specifically, adversaries are viewed as \textit{inferior positives} that induce weaker learning signals, or as \textit{hard negatives} exhibiting higher contrast to other negative samples. In the asymmetric fashion, the adverse impacts of conflicting objectives between CL and adversarial learning can be effectively mitigated. Experiments show that our approach consistently outperforms existing Adversarial CL methods across different finetuning schemes. The proposed A-InfoNCE is also a generic form that can be readily extended to other CL methods. Code is available at \url{https://github.com/yqy2001/A-InfoNCE}.


\end{abstract}

\section{Introduction}

Well-performed models trained on clean data can suffer miserably when exposed to simply-crafted adversarial samples~\cite{szegedy2014intriguing,goodfellow2014explaining,carlini2017towards,dong2018boosting}. 
There has been many adversarial defense mechanisms designed to boost model robustness using labeled data~\cite{kannan2018adversarial,shafahi2019adversarial,zhang2019you,wong2020fast,zhang2019theoretically,zhu2019freelb,athalye2018obfuscated}. In practice, however, obtaining large-scale annotated data can be far more difficult and costly than acquiring unlabeled data. Leveraging easily-acquired unlabeled data for adversarial learning, thus becomes particularly attractive.

Contrastive Learning (CL)~\cite{hadsell2006dimensionality}, which performs instance discrimination~\cite{wu2018unsupervised} (Figure 1 (a)) by maximizing agreement between augmentations of the same instance in the learned latent features while minimizing the agreement between different instances,
has made encouraging progress in self-supervised learning~\cite{pmlr-v119-chen20j,he2020momentum,chen2020improved,grill2020bootstrap}. Due to its effectiveness in learning rich representations and competitive performance over fully-supervised methods, CL has seen a surge of research in recent years, such as
positive sampling~\cite{pmlr-v119-chen20j,tian2020contrastive,bachman2019learning,tian2020makes}, negative sampling~\cite{he2020momentum,kalantidis2020hard,chuang2020debiased,wu2018unsupervised}, 
pair reweighting~\cite{chuang2020debiased,robinson2020contrastive}, and different contrast methods~\cite{grill2020bootstrap,caron2020unsupervised,li2020prototypical}.

\begin{figure}[t]
\centering
\includegraphics[scale=0.43]{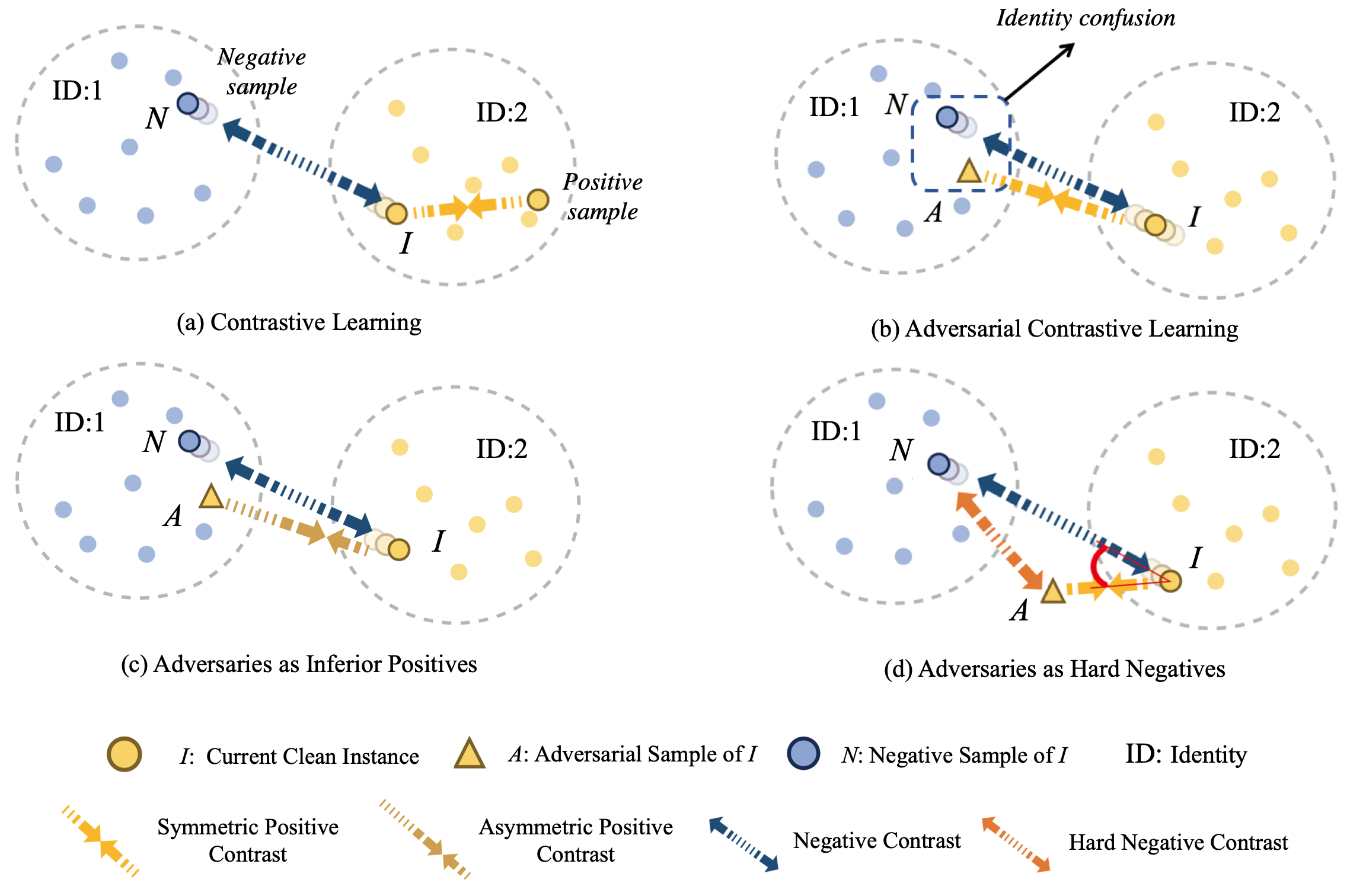}
\caption{
Illustrations of ($a$) Contrastive Learning; ($b$) Adversarial Contrastive Learning; and our proposed methods for viewing adversarial samples asymmetrically as: ($c$) Inferior Positives (asymmetric contrast), and ($d$) Hard Negatives. In each circle, data points are augmentations of the same instance, sharing the same \textit{Identity}. In ($b$), the Adversarial sample (\textit{A}) shares the same Identity (\textit{ID:2}) as the current Instance (\textit{I}), but resides close to a different Identity (\textit{ID:1}), thus \textit{Identity Confusion} problem occurs. Specifically, the Adversarial sample (\textit{A}) of Instance \textit{(I)} exhibits similar representations to the Negative sample (\textit{N}) of (\textit{I}), which makes the positive contrast (\textit{A}$\leftrightarrow$\textit{I}) and negative contrast (\textit{N}$\leftrightarrow$\textit{I}) undermine each other in the training process (\textit{colored figure}). 
}
\label{fig:intro}
\end{figure}

Recently, contrastive learning has been extended to adversarial learning tasks in a self-supervised manner, leading to a new area of \textit{adversarial contrastive learning} (Adversarial CL) ~\cite{kim2020adversarial,fan2021does,jiang2020robust,gowal2020self}. 
The main idea is to generate adversarial samples as additional positives of the same instance~\cite{kim2020adversarial,fan2021does,jiang2020robust} for instance-wise attack, and maximize the similarity between clean views of the instance and their adversarial counterparts as in CL, while also solving the min-max optimization problem following canonical adversarial learning objective~\cite{madry2018towards,shafahi2019adversarial,zhang2019you,wong2020fast,zhang2019theoretically,zhu2019freelb}.
For example, RoCL\cite{kim2020adversarial} first proposed an attack mechanism against contrastive loss to confuse the model on instance-level identity, in a self-supervised adversarial training framework. AdvCL\cite{fan2021does} proposed to minimize the gap between unlabeled contrast and labeled finetuning by introducing pseudo-supervision in the pre-training stage.


Although these Adversarial CL methods showed improvement on model robustness, we observe that a direct
extension from CL to adversarial learning (AL) can introduce ineffective CL updates during training.
The core problem lies in that they add worst-case perturbations $\delta$ that no longer guarantee the preservation of instance-level identity~\cite{kim2020adversarial} (\textit{i.e.}, different from other data augmentation methods,  adversarial samples can reside faraway from the current instance in the feature space after several attack iterations, because the attack objective is to make adversaries away from the current instance while approximating other instances, against the CL objective). 
As illustrated in Figure~\ref{fig:intro}(b), when the adversarial sample (\textit{A}) of the current instance (\textit{I}) are in close proximity to negative samples (\textit{N}), CL objective minimizes the agreement between negative samples and current instance (\textit{I} and \textit{N} are pushed away from each other), while AL objective maximizes the agreement between adversarial samples and current instance (\textit{A} and \textit{I} are pulled together as \textit{A} is considered as an augmented view of \textit{I}). Meanwhile, \textit{A} and \textit{N} share similar representations, which renders the two objectives contradicting to each other. We term this conflict as ``\textit{identity confusion}'', it means $A$ attracts and `confuses' $I$ with a false identity induced by $N$, which impedes both CL and AL from achieving their respective best performance.

To address this issue of \textit{identity confusion}, we propose to treat adversarial samples unequally and discriminatingly, and design a generic asymmetric InfoNCE objective (\textit{A-InfoNCE}), in order to model the asymmetric contrast strengths between positive/negative samples.
Firstly, to mitigate the direct pull between adversarial sample (\textit{A}) and current instance (\textit{I}) (Figure~\ref{fig:intro} (c)) that might dampen the effectiveness of CL, we propose to treat adversarial samples as \textit{inferior positives} that induce weaker learning signals to attract their counterparts in a lower degree when performing positive contrasts.
This asymmetric consideration in AL promises a trade-off and reduces conflicting impact on the CL loss.

Secondly, to encourage adversarial samples (\textit{A}) to escape from false identities induced by negative samples (\textit{N}) that share similar representations to (\textit{A}) (pushing \textit{A} away from \textit{N}) (Figure~\ref{fig:intro}(d)), we consider adversarial samples (\textit{A}) as \textit{hard negatives}~\cite{robinson2020contrastive} of other negative samples (\textit{N}), by strengthening the negative contrast between \textit{A} and \textit{N} in CL computation.
To effectively sample true adversarial negatives and re-weight each sample, we follow positive-unlabeled learning~\cite{du2014analysis,elkan2008learning} and contrastive negatives reweighting~\cite{robinson2020contrastive,chuang2020debiased} practice. 

Our contributions are summarized as follows: 
$1)$
  We propose an generic asymmetric InfoNCE loss, \textit{A-InfoNCE}, to address the \textit{identity confusion} problem in Adversarial CL, by viewing adversarial samples 
     as \textit{inferior positives} or \textit{hard negatives}.
  2) Our approach is compatible to existing Adversarial CL methods, by simply replacing standard CL loss with \textit{A-InfoNCE}. 
   3) Experiments on CIFAR-10, CIFAR-100 and STL-10 show that our approach consistently outperforms existing Adversarial CL methods.

\section{Asymmetric InfoNCE}

\subsection{Notations}


\paragraph{Contrastive Learning (CL) }

CL aims to learn generalizable features by maximizing agreement between self-created positive samples while contrasting to negative samples.
In typical contrastive learning, each instance $x$ will be randomly transformed into two views $(x_1, x_2)$, then fed into a feature encoder $f$ with parameters $\theta$ to acquire normalized projected features, \textit{i.e.}, $z_i = f(x_i;\theta)$.
Let $\mathcal{P}(i)$ denote the set of positive views of $x_i$, containing the views transformed from $x$ with the same instance-level \textit{identity} (\textit{e.g.}, augmentations of the original image $x_i$); $\mathcal{N}(i)$ denotes the set of negative views of $x_i$, containing all the views from other instances.
The conventional InfoNCE loss function~\cite{oord2018representation} used in CL for a positive pair $(x_i,x_j)$ is defined as:
\begin{align}
    \mathcal{L}_{\rm CL}(x_i,x_j) 
    =
    - \log \frac
    {\exp({\rm sim}(z_i, z_j)/t)}
    {
        \exp({\rm sim}(z_i, z_j)/t) + 
        \sum_{k\in \mathcal{N}(i)} \exp({\rm sim}(z_i, z_k)/t)
    }
\end{align}
where $x_i$ serves as the anchor, ${\rm sim}(z_i, z_j)$ denotes a similarity metric (\textit{e.g.}, cosine similarity) between $z_i$ and $z_j$, and $t$ is a temperature parameter.
The final loss of the CL problem is averaged over all positive pairs of instances.

\paragraph{Adversarial CL}
Adversarial CL can be regarded as an extension of CL by adding adversarial samples into the positive sets $\mathcal{P}(\cdot)$ to contrast. Adversarial CL is typically modeled as the following min-max optimization formulation to incorporate instance-wise attack~\cite{madry2018towards,fan2021does}:
\begin{align}
    \min_\theta \mathbb{E}_{x\in \mathcal{X}} \max_{||\delta||_\infty \leq \epsilon} \sum_i\sum_{j\in \mathcal{P}(i)}
    \mathcal{L}_{\rm CL}(x_i, x_j),\quad \mathcal{P}(i)\leftarrow \mathcal{P}(i) \cup \{\hat{x}_i+\delta\}
\end{align}
where $\hat{x}_i$ is the view of $x_i$ used to generate adversarial samples, $\delta$ is the adversarial perturbation whose infinity norm is constrained as less than $\epsilon$.
In the above formulation, the inner maximization problem constructs adversarial samples by maximizing the contrastive loss, and the outer minimization problem optimizes the expected worst-case loss w.r.t. the feature encoder $f$. 

\subsection{Asymmetric InfoNCE: A Generic Learning Objective}
Current Adversarial CL frameworks directly inherit CL's conventional contrastive loss (\textit{e.g.}, InfoNCE) to evaluate the similarity between adversarial and clean views in a symmetric fashion. This can result in ineffective or even conflicting updates during CL training as aforementioned.
To address this challenge, we propose a generic Asymmetric InfoNCE loss (\textit{A-InfoNCE}) to incorporate the asymmetric influences between different contrast instances, given by:
\begin{equation} \label{eq:1}
\resizebox{\textwidth}{!}{
    $\mathcal{L}^{\rm asym}_{\rm CL} (x_i, x_j;{\alpha},\lambda^p, \lambda^n) 
    = 
    - \log \frac
    {\lambda^p_j \cdot \exp({\rm sim^{\alpha}}(z_i, z_j)/t)}
    {\lambda^p_j \cdot \exp({\rm sim^{\alpha}}(z_i, z_j)/t) + {\sum_{k\in \mathcal{N}(i)} \lambda^n_k\cdot\exp({\rm sim^{\alpha}}(z_i, z_k)/t)}}
    $
}
\end{equation}
where $\rm sim^{\alpha}(\cdot)$ is a generalized similarity metric that enables the incorporation of asymmetric relationships (a concrete instantiation is described in the next section); $\lambda^p$ and $\lambda^n$ are asymmetric weighting factors for positive and negative pairs, respectively.
It is worth noting that although A-InfoNCE is proposed to address the \textit{identity confusion} issue in Adversarial CL, it can be easily extended to other CL settings when the asymmetric characteristics between different views need to be captured.
A-InfoNCE can also generalized to many existing CL methods, for example, $\mathcal{P}(i)$ and $\mathcal{N}(i)$ can be altered to different choices of positive and negative views; ${\rm sim^{\alpha}}(z_i, z_j)$ is also changeable to a symmetric similarity metric for $z_i$ and $z_j$. $\lambda^p$ and $\lambda^n$ control the weights of different positive/negative pairs. Generalization strategies are itemized below:
\begin{itemize}
    \item If ${\rm sim^{\alpha}}(z_i, z_j)$ is a symmetric similarity metric and $\lambda^p, \lambda^n = 1$, it degrades to the conventional InfoNCE loss used in CL~\cite{pmlr-v119-chen20j}.
    \item If $\mathcal{P}(i)$ is altered, it corresponds to positives sampling~\cite{tian2020contrastive,bachman2019learning,tian2020makes}
    . When we add adversaries into $\mathcal{P}(i)$, it degenerates to the conventional Adversarial CL objectives, where $\lambda^p, \lambda^n = 1$ with symmetric ${\rm sim^{\alpha}}(z_i, z_j)$~\cite{kim2020adversarial,jiang2020robust,fan2021does}.
    \item If we seek better $\mathcal{N}(i)$, it echos negative sampling methods~\cite{robinson2020contrastive,kalantidis2020hard} such as Moco~\cite{he2020momentum}, which maintains a queue of consistent negatives; or mimics DCL~\cite{chuang2020debiased} that debiases $\mathcal{N}(i)$ into true negatives. 
    \item If we change $\lambda^p$ and $\lambda^n$, it mirrors the pair reweighting works~\cite{chuang2020debiased,robinson2020contrastive} that assign different weights to each pair according to a heuristic measure of importance such as similarity.
\end{itemize}
While most existing methods adopt a symmetric similarity metric, we claim that in some scenarios the asymmetric similarity perspective needs to be taken into account, especially when the quality and property of different views vary significantly.
In this paper, we focus on the study of Adversarial CL, and demonstrate the benefits of 
capturing the asymmetric relationships between adversaries and clean views.
Specifically, we design two instantiations to model the asymmetric relationships between adversarial and clean samples, as detailed in next section.
Both instantiations 
can be integrated into the proposed \textit{A-InfoNCE} framework.






\section{Adversarial Asymmetric Contrastive Learning}


This section explains the instantiations of the \textit{A-InfoNCE} loss for Adversarial CL. From the \textit{inferior-positive} perspective, to reduce the impact of identity confusion, we first design a new asymmetric similarity metric ${\rm sim^{\alpha}}(z_i, z_j^{adv})$ for modeling the asymmetric relationships and weakening the learning signals from adversarial examples. From the \textit{hard-negative} perspective, we view adversaries as hard negatives for other negative samples, and reweight each negative pairs by assigning similarity-dependent weights to ease the identity confusion. 


\subsection{Adversarial Samples as Inferior Positives}


Adversarial samples with different identities may attract their anchors (clean samples) in a contradicting manner to the exertion of CL. By weakening the learning signal from these adversarial examples in positive contrast (as \textit{inferior positives} that attract the anchors less), we can effectively mitigate the undesired pull from clean samples via an adaptive gradient stopping strategy.

\subsubsection{Asymmetric Similarity Function.}

As the symmetric nature of InfoNCE can bring conflicts in Adversarial CL, we design a new asymmetric similarity function ${\rm sim^{\alpha}}(z_i, z_j)$ for \textit{A-InfoNCE}, by manipulating the scale of gradient for each contrasted branch. We decompose it into two parts for each branch:
\begin{align} \label{eq:2}
    {\rm sim^{\alpha}}(z_i, z_j) 
    = 
    \alpha \cdot {\rm \overline{sim}}(z_i, z_j) 
    +
    (1 - \alpha) \cdot {\rm \overline{sim}}(z_j, z_i)
\end{align}
where ${\overline{\rm sim}(a, b)}$ means the one-sided similarity of $a$ to $b$, \textit{i.e.}, when maximizing ${\overline{\rm sim}(a, b)}$, we freeze $b$ and only move $a$ towards $b$. This can be implemented by stopping the gradient back-propagation for $b$ and only optimizing $a$.

We use a hyperparameter $\alpha$ to control how much $z_i$ and $z_j$ head towards each other. For a clean sample and an adversarial sample, we let $\alpha$ denote the coefficient of the clean branch's movement. If $\alpha$ is 0, it performs total gradient freezing on the clean branch and only adversarial representations are optimized through training. Our empirical analysis finds that $\alpha$ is relatively easy to tune for boosted performance. We show that any value lower than 0.5 brings reasonable performance boost (see Figure~\ref{fig:ablation1}), when clean samples move less towards adversaries, following the intrinsic asymmetric property of Adversarial CL.

\subsubsection{Adaptive $\alpha$-annealing.}


When the \textit{identity confusion} is at play, it is necessary to treat adversarial samples inferior to ensure model robustness. But as training progresses, when model learns robust representations and the negative identity-changing impact of adversarial perturbation wanes, we consider adversarial perturbation as strong augmentations, equal to other typical transformations~\cite{pmlr-v119-chen20j}. 

The question is how to measure the reduction of instance confusion effect. Here we take a geometry perspective and propose to adaptively tune the proportional coefficient $\alpha$ on-the-fly based on Euclidean distance. Let $d_{i,j} = ||z_i - z_j||_2$ denote the distance between an original image and its adversarial view in the representation space.
Given $\alpha_{min}$, $d_{max}$, $\alpha_{max}$, $d_{min}$,
the goal is for $\alpha$ to be $\alpha_{max}$ when the distance approximates $d_{min}$, and $\alpha_{min}$ to be close to $d_{max}$. During training, we first compute the current representation distance $d$, then use a simple linear annealing strategy to compute $\alpha$:
\begin{align}
    \alpha = \alpha_{min} + (d_{max}-d)\frac{\alpha_{max}-\alpha_{min}}{d_{max}-d_{min}}
\end{align}
$d_{min}$ and $\alpha_{min}$ can be treated as hyperparameters. $\alpha_{max}$ is 0.5, indicating adversarial perturbation is equal to other transformations and ${\rm sim}^\alpha(z_i, z_j)$ degrades to the symmetric similarity. Moreover, we use the first $N$ epochs as a warm-up to compute the average distance as $d_{max}$, in which period $\alpha$ is fixed.

\subsubsection{Adversarial CL Loss with Inferior Positives.}
With the above asymmetric similarity function $\rm sim^\alpha(\cdot)$ and the \textit{A-InfoNCE} loss function $\mathcal{L}^{\rm asym}_{\rm CL}(x_i, x_j;{\alpha},\lambda^p, \lambda^n)$, the complete Adversarial CL loss with \textit{inferior positives} (IP) can be written as: 
\begin{small}
\begin{align} \label{eq:ip}
        \mathcal{L}^{\rm IP} 
    = 
        \sum_i\sum_{j\in \mathcal{P}(i)} \mathcal{L}^{\rm asym}_{\rm CL}(x_i, x_j; 0.5, 1, 1) 
    +
        \gamma \cdot \sum_i\sum_{j\in \mathcal{P}(i)} \mathcal{L}^{\rm asym}_{\rm CL}(x_i, x_j^{adv}; \alpha, 1, 1)
\end{align}
\end{small}
where the first part stands for standard CL loss that maximizes the similarity between two clean views, which is symmetric ($\alpha=0.5$) with $\lambda^p = \lambda^n = 1$, degrading to the conventional InfoNCE loss. The second part is a robust CL loss that maximizes the agreement between clean and adversarial views, but uses the asymmetric similarity function (\ref{eq:2}) with a hyperparameter $\alpha$ that gives weaker learning signals to the counterparts of inferior adversarial samples. The hyperparameter $\gamma$ balances the robustness and accuracy objectives.

\subsection{Adversarial Samples as Hard Negatives}

Besides inferior positives, we also propose an alternative view of adversaries as \textit{hard negatives}~\cite{robinson2020contrastive} that be pushed away from surrounding data points with higher weights. This can potentially assuage the confusion brought by adversarial samples of the current instance residing too close to the negative samples of the same instance (as illustrated in Figure 1 (d)). Furthermore, this strategy encourages the model towards more robustness-aware, by giving adversarial samples possessing undiscriminating features higher weights in the pretraining stage, further enhancing Adversarial CL.


In practice, we assign a weight of similarity to each pair. To set a basis for weight assigning, we adopt a simple and adaptive weighting strategy used in~\cite{robinson2020contrastive}, \textit{i.e.}, taking each pair's similarity as its weight, with $w_{i,j} = \exp({\rm {sim}}(z_i, z_j)/t)$. By doing so, the adversaries with bad instance-level identity (greater similarity to negative samples) can be automatically assigned with higher weights. The weights can adaptively decay as the instance identity recovers during training.

However, as the commonly-used $\mathcal{N}(i)$ is uniformly sampled from the entire data distribution $p(x)$~\cite{chuang2020debiased} (\textit{e.g.}, SimCLR~\cite{pmlr-v119-chen20j} uses other instances in the current batch as negative samples), simply taking similarities as weights may heavily repel semantically-similar instances whose embeddings should be close. To estimate the true negatives distribution $p^-(x)$ , we take advantage of PU-learning~\cite{du2014analysis,elkan2008learning} and resort to DCL,HCL~\cite{chuang2020debiased,robinson2020contrastive} to debias negative sampling.

PU-learning~\cite{du2014analysis} decomposes the data distribution as: $p(x) = \tau p^+ (x) + (1-\tau) p^- (x)$, where $p^+(x), p^-(x)$ denote the distribution of data from the same or different class of $x$, and $\tau$ is the class prior. Thus
$p^{-}(x)$ can be rearranged as $p^{-}(x)=\big(p(x) - \tau p^+ (x)\big)/(1-\tau)$. We can use all instances and positive augmentations containing adversarial samples of $x$ to estimate $p(x)$ and $p^+(x)$, respectively. Following~\cite{chuang2020debiased}, we debias the negative contrast part in (\ref{eq:1}) as:
\begin{small}
\begin{align}
    \frac{1}{1-\tau}
    \Big(
        \sum_{k\in \mathcal{N}(i)} w_{i,k}^n \cdot \exp({\rm sim^{\alpha}}(z_i, z_k)/t)
    -
        \frac{N}{M} \cdot \tau 
        \sum_{j\in \mathcal{P}(i)} w_{i,j}^p \cdot \exp({\rm sim^{\alpha}}(z_i, z_j)/t)
    \Big)
\end{align}
\end{small}
where $M, N$ are the numbers of postives and negatives, $w_{i,k}^n$ is the aforementioned weights for negatives, $w_{i,j}^p$ is a expandable weight for positives (set as 1 in our implementation, other choices can be further explored in the future work).

\subsubsection{Adversarial CL Loss with Hard Negatives.}
We substitute (7) into the \textit{A-InfoNCE} loss function (\ref{eq:1}) and rearrange it, acquiring the instantiation of \textit{A-InfoNCE} loss with \textit{hard negatives} (HN), with concrete forms of $\lambda^p$ and $\lambda^n$ as:
\begin{small}
\begin{align} \label{eq:hn}
        \mathcal{L}^{HN} 
    = 
        \sum_i\sum_{j\in \mathcal{P}(i)}
        \mathcal{L}^{\rm asym}_{\rm CL}(x_i, x_j; \alpha, \frac{M-(M+N)\tau}{M-M\tau}w_{i,j}^p, \frac{1}{1-\tau}w_{i,k}^n),\quad k\in\mathcal{N}(i)
\end{align}
\end{small}
Due to the lack of class information, we treat $\tau$ as a hyperparameter and set as~\cite{chuang2020debiased} suggested. 

\subsubsection{Combined Adversarial CL Loss.}
Finally, we can view adversaries both as inferior positives and hard negatives for other negative samples. This leads to following combined Adversarial CL loss:
\begin{small}
\begin{align} \label{eq:ip+hn}
        \mathcal{L}^{IP+HN}
    &=
        \sum_i\sum_{j\in \mathcal{P}(i)} \mathcal{L}^{\rm asym}_{\rm CL}(x_i, x_j; 0.5, \frac{M-(M+N)\tau}{M-M\tau}w_{i,j}^p, \frac{1}{1-\tau}w_{i,k}^n) \ 
    + \nonumber \\
        & \gamma \cdot \sum_i\sum_{j\in \mathcal{P}(i)} \mathcal{L}^{\rm asym}_{\rm CL}(x_i, x_j^{adv}; \alpha, \frac{M-(M+N)\tau}{M-M\tau}w_{i,j}^p, \frac{1}{1-\tau}w_{i,k}^n),\quad k\in\mathcal{N}(i)
\end{align}
\end{small}

\section{Experiments}

To demonstrate the effectiveness and generalizability of the proposed approach, we present experimental results across different datasets and model training strategies. Our methods are compatible with existing Adversarial CL frameworks, and can be easily incorporated by replacing their CL loss. 
We choose two baselines and replace their loss with $\mathcal{L}^{IP}$(in Equation~\ref{eq:ip}), $\mathcal{L}^{HN}$(\ref{eq:hn}) and $\mathcal{L}^{IP+HN}$(\ref{eq:ip+hn}) for evaluation.

\textbf{Datasets.} We mainly use CIFAR-10 and CIFAR-100 for our experiments. Each dataset has 50,000 images for training and 10,000 for test. STL-10 is also used for transferability experiments. Following previous work~\cite{fan2021does}, we use ResNet-18~\cite{he2016deep} as the encoder architecture in all experiments.

\setlength{\tabcolsep}{0pt}
\setlength{\arrayrulewidth}{0.2mm}
\renewcommand{\arraystretch}{1.3}
\begin{table}[t] %
\caption{Results for replacing the objectives of the two baselines with $\mathcal{L}^{IP}$, $\mathcal{L}^{HN}$ and $\mathcal{L}^{IP+HN}$, in Standard Accuracy (SA) and Robust Accuracy (RA).
The pre-trained methods are evaluated under the Linear Probing (LP), Adversarial Linear Finetuning (ALF) and Adversarial Full Finetuning (AFF) strategies.
Supervised methods are trained under conventional adversarial training scheme
}
\centering
\fontsize{8.5}{9}\selectfont 
\scalebox{0.9}{
\begin{tabularx}{1\textwidth}
{  m{1.2cm}
   m{1.6cm}
   m{2.2cm}
    P{1.1cm}
    P{1.2cm}
    P{1.1cm}
    P{1.3cm}
    P{1.1cm}
    P{1.1cm}
    P{0.5cm}
    }
\toprule
\multicolumn{1}{l}{\multirow{3}{*}{Dataset}} & \multicolumn{2}{c}{\multirow{3}{*}{\makecell[c]{Pre-training \\ Methods}}} & \multicolumn{6}{c}{Finetuning Strategies} \\ \cmidrule(l){4-9} 
\multicolumn{1}{l}{} & \multicolumn{2}{c}{} & \multicolumn{2}{c}{Linear Probing} & \multicolumn{2}{l}{\begin{tabular}[c]{@{}c@{}}Adversarial Linear\\ Finetuning\end{tabular}} & \multicolumn{2}{c}{\begin{tabular}[c]{@{}c@{}}Adversarial Full\\ Finetuning\end{tabular}} \\ \cmidrule(l){4-9} 
\multicolumn{1}{c}{} & \multicolumn{2}{c}{} & \multicolumn{1}{c}{SA} & \multicolumn{1}{c}{RA} & \multicolumn{1}{c}{SA} & \multicolumn{1}{c}{RA} & \multicolumn{1}{c}{SA} & \multicolumn{1}{c}{RA} \\
\hline\hline
\multirow{10}{*}{\makecell[c]{CIFAR \\ 10}} & \multirow{2}{*}{Supervised} & AT~\cite{madry2018towards} &-&-&-&-& 78.99 & 47.41 & \scriptsize{1}\\
 &  & TRADES~\cite{zhang2019theoretically} &-&-&-&-& 81.00 & 53.27 & \scriptsize{2} \\ \cmidrule(l){2-9} 
 & \multirow{8}{*}{\begin{tabular}[c]{@{}c@{}}Self-\\ Supervised \end{tabular}} & RoCL~\cite{kim2020adversarial} & 83.84 & 38.98 & 79.23 & 47.82 & 77.83 & 50.54 & \scriptsize{3} \\
 &  & \   \ w/ $\mathcal{L}^{IP}$ & \textbf{87.63} & 41.46 & \textbf{84.15} & 50.08 & 78.97 & 50.29 & \scriptsize{4} \\
 &  & \   \ w/ $\mathcal{L}^{HN}$ & 84.14 & 40.00 & 79.40 & 48.31 & 78.84 & 51.73 & \scriptsize{5} \\
 &  & \   \ w/ $\mathcal{L}^{IP+HN}$ & 85.69 & \textbf{42.96} & 81.91 & \textbf{50.90} & \textbf{80.06} & \textbf{52.95} & \scriptsize{6} \\ \cmidrule(l){3-9}
 &  & AdvCL~\cite{fan2021does} & 81.35 & 51.00 & 79.24 & 52.38 & 83.67 & 53.35 & \scriptsize{7} \\
 &  & \   \ w/ $\mathcal{L}^{IP}$ & 82.37 & 52.33 & 80.05 & \textbf{53.22} & \textbf{84.12} & 53.56 & \scriptsize{8} \\
 &  & \   \ w/ $\mathcal{L}^{HN}$ & 81.34 & 52.61 & 78.69 & 53.20 & 83.44 & \textbf{54.07} & \scriptsize{9} \\
 &  & \   \ w/ $\mathcal{L}^{IP+HN}$ & \textbf{83.15} & \textbf{52.65} & \textbf{80.41} & 53.19 & 83.93 & 53.74 & \scriptsize{10} \\ \midrule
\multirow{10}{*}{\makecell[c]{CIFAR \\ 100}} & \multirow{2}{*}{Supervised} & AT~\cite{madry2018towards} & -&-&-&- & 49.49 & 23.00 & \scriptsize{11} \\
 &  & TRADES~\cite{zhang2019theoretically} & -&-&-&- & 54.59 & 28.43 & \scriptsize{12} \\ \cmidrule(l){2-9} 
 & \multirow{8}{*}{\begin{tabular}[c]{@{}c@{}}Self-\\ Supervised\end{tabular}} & RoCL~\cite{kim2020adversarial} & 55.71 & 18.49 & 49.30 & 25.84 & 51.19 & 26.69 & \scriptsize{13} \\
 &  & \   \ w/ $\mathcal{L}^{IP}$ & 59.30 & 21.34 & 54.49 & \textbf{30.33} & 52.39 & 27.84 & \scriptsize{14} \\
 &  & \   \ w/ $\mathcal{L}^{HN}$ & 58.77 & 21.17 & 56.38 & 28.03 & \textbf{55.85} & 29.57 & \scriptsize{15} \\
 &  & \   \ w/ $\mathcal{L}^{IP+HN}$ & \textbf{59.74} & \textbf{22.54} & \textbf{57.57} & 29.22 & 55.79 & \textbf{29.92} & \scriptsize{16} \\ \cmidrule(l){3-9}
 &  & AdvCL~\cite{fan2021does} & 47.98 & 27.99 & \textbf{47.45} & 28.29 & 57.87 & 29.48 & \scriptsize{17} \\ 
 &  & \   \ w/ $\mathcal{L}^{IP}$ & 49.48 & 28.84 & 45.39 & 28.40 & \textbf{59.44} & 30.49 & \scriptsize{18} \\
 &  & \   \ w/ $\mathcal{L}^{HN}$ & 49.44 & 29.01 & 47.32 & \textbf{28.69} & 58.41 & 29.93 & \scriptsize{19} \\
 &  & \   \ w/ $\mathcal{L}^{IP+HN}$ & \textbf{50.59} & \textbf{29.12} & 45.72 & 28.45 & 58.70 & \textbf{30.66} & \scriptsize{20} \\ \bottomrule
\end{tabularx}}
\label{table:1}
\end{table}

\textbf{Baselines.} We compare with two baselines: RoCL~\cite{kim2020adversarial}, the first method to combine CL and AL; and AdvCL~\cite{fan2021does}, the current state-of-the-art framework. During experiments, we observe severe overfitting of AdvCL when training 1000 epochs (experiment setting in the original paper), with performance inferior to training for 400 epochs.
Thus, we pre-train 400 epochs on AdvCL at its best-performance setting. All other settings are the same as original papers except for some hyperparameter tuning. Our methods are also compatible with some recent work like SwARo~\cite{wahed2022adversarial} and CLAF~\cite{rahamim2022robustness}, by modeling the asymmetry between clean and adversarial views as aforementioned.

\textbf{Evaluation.}
Following \cite{jiang2020robust} and \cite{fan2021does}, we adopt three finetuning strategies to evaluate the effectiveness of contrastive pre-training: $1)$ Linear Probing (LP): fix the encoder and train the linear classifier; $2)$ Adversarial Linear Finetuning (ALF): adversarially train the linear classifier; $3)$ Adversarial Full Finetuning (AFF): adversarially train the full model. We consider two evaluation metrics: $1)$ Standard Accuracy (SA): classification accuracy over clean images; $2)$ Robust Accuracy (RA): classification accuracy over adversaries via PGD-20 attacks~\cite{madry2018towards}. 
Robustness evaluation under more diverse attacks is provided in the appendix.


\subsection{Main Results}
In Table \ref{table:1}, we report standard accuracy and robust accuracy of each model, learned by different pre-training methods over CIFAR-10 and CIFAR-100. Following previous works~\cite{kim2020adversarial,jiang2020robust,fan2021does} and common practice in contrastive learning~\cite{pmlr-v119-chen20j,he2020momentum}, we first use unlabeled images in CIFAR-10/-100 to pre-train, then introduce labels to finetune the model. As shown in Table \ref{table:1}, our methods achieve noticeable performance improvement over baselines in almost all scenarios, when replacing the original loss with our proposed adversarial CL loss.

In comparison with RoCL, $\mathcal{L}^{IP}$ brings significant performance boost on both standard and robust accuracy consistently across different training methods (row 4 vs. 3, row 14 vs. 13) (except for RA of AFF on CIFAR10).
Comparing to AdvCL,  $\mathcal{L}^{IP}$  also brings noticeable margin (row 8 vs. 7, row 18 vs. 17). This can be attributed to that $\mathcal{L}^{IP}$ aims to lower the priority of adversaries and prevent clean samples moving towards other instances, which results in better instance discrimination and improves clean~\cite{wu2018unsupervised} and robust accuracy.
$\mathcal{L}^{HN}$ also yields substantial boost on robust and standard accuracy (\textit{e.g.}, row 15 vs. 13). 
We hypothesize this is due to that  $\mathcal{L}^{HN}$ helps
alert the model to adversarial samples by assigning higher weights for adversaries in negative contrast. 
When combined together, in most settings both standard and robust accuracy are further boosted, especially for Linear Probing. This is because directly mitigating the negative impact of \textit{identity confusion} by $\mathcal{L}^{IP}$ and helping adversarial get rid of false identities by $\mathcal{L}^{HN}$ can complement each other, bringing further performance boost.
\subsection{Transferring Robust Features}

\setlength{\tabcolsep}{0pt}
\setlength{\arrayrulewidth}{0.2mm}
\renewcommand{\arraystretch}{1.3}
\begin{table}[t]
\fontsize{8.5}{10}\selectfont  
\centering
\caption{Transferring results from CIFAR-10/100 to STL-10, compared with AdvCL~\cite{fan2021does}, evaluated in Standard accuracy (SA) and Robust accuracy (RA) across different finetuning methods with ResNet-18
}
\scalebox{0.9}{
\begin{tabularx}{1\textwidth}
{
   P{2cm}
   m{2.4cm}
    P{1.2cm}
    P{1.3cm}
    P{1.2cm}
    P{1.4cm}
    P{1.2cm}
    P{1.3cm}
}
\toprule
\multirow{3}{*}{Dataset} & \multirow{3}{*}{\makecell[c]{Pre-training \\ Methods}} & \multicolumn{6}{c}{Finetuning Strategies} \\ \cline{3-8} 
 &  & \multicolumn{2}{c}{Linear Probing} & \multicolumn{2}{c}{\makecell[c]{Adversarial Linear \\ Finetuning}} & \multicolumn{2}{c}{\makecell[c]{Adversarial Full \\ Finetuning}} \\ \cline{3-8} 
 &  & SA & RA & SA & RA & SA & RA \\
 \hline\hline
\multirow{4}{*}{\makecell[c]{CIFAR10\\$\downarrow$\\STL10}} & AdvCL~\cite{fan2021does} & 64.45 & 37.25 & 60.86 & 38.84 & 67.89 & 38.78 \\
 & \   \ w/ $\mathcal{L}^{IP}$ & 64.83 & 37.30 & 61.95 & 38.90 & \textbf{68.25} & 39.03 \\
 & \   \  w/ $\mathcal{L}^{HN}$ & 65.24 & \textbf{38.18} & \textbf{62.83} & \textbf{39.70} & 67.88 & \textbf{39.75} \\
 & \   \  w/ $\mathcal{L}^{IP+HN}$ & \textbf{67.19} & 37.00 & 61.34 & 39.35 & 67.95 & 39.12 \\ \hline
\multirow{4}{*}{\makecell[c]{CIFAR100\\$\downarrow$\\STL10}} & AdvCL~\cite{fan2021does} & 52.28 & 30.01 & 49.84 & 32.14 & 63.13 & 35.24 \\
 & \   \ w/ $\mathcal{L}^{IP}$ & 52.65 & \textbf{31.33} & 50.18 & 33.15 & 63.26 & \textbf{35.34} \\
 & \   \ w/ $\mathcal{L}^{HN}$ & 51.88 & 31.29 & \textbf{50.73} & \textbf{33.62} & 62.91 & 34.88 \\
 & \   \ w/ $\mathcal{L}^{IP+HN}$ & \textbf{53.41} & 31.30 & 51.10 & 33.23 & \textbf{63.69} & 35.09 \\ \bottomrule
\end{tabularx}}
\label{table:2}
\end{table}

Learning robust features that are transferable is a main goal in self-supervised adversarial learning. It is of great significance if models pre-trained with a huge amount of unlabeled data possess good transferability by merely light-weight finetuning. For example, Linear Probing is often 10$\times$ quicker than conventional adversarial training, with only a linear classifier trained.

Here we evaluate the robust transferability of the proposed approach, by transfering CIFAR-10 and CIFAR-100 to STL-10, \textit{i.e.}, use unlabeled images in CIFAR-10/-100 to pretrain, then use STL-10 to finetune and evaluate the learned models. As shown in Table~\ref{table:2}, our methods yield both clean and robust accuracy gains in most settings, up to 1.48\% (33.62\% vs. 32.14\%) in robust accuracy and 2.74\% (67.19\% vs. 64.45\%) in clean accuracy.

\subsection{Ablation studies}

We design a basic adversarial contrastive model, named CoreACL, to study the effect of each component in our proposed methods. 
CoreACL only contains the contrastive component with three positive views: two clean augmented views and one adversarial view of the original image. 

\subsubsection{Fixed $\alpha$ for Asymmetric Similarity Function.}
\begin{wrapfigure}[14]{r}{8.0cm}
    \centering
    \includegraphics[scale=0.27]{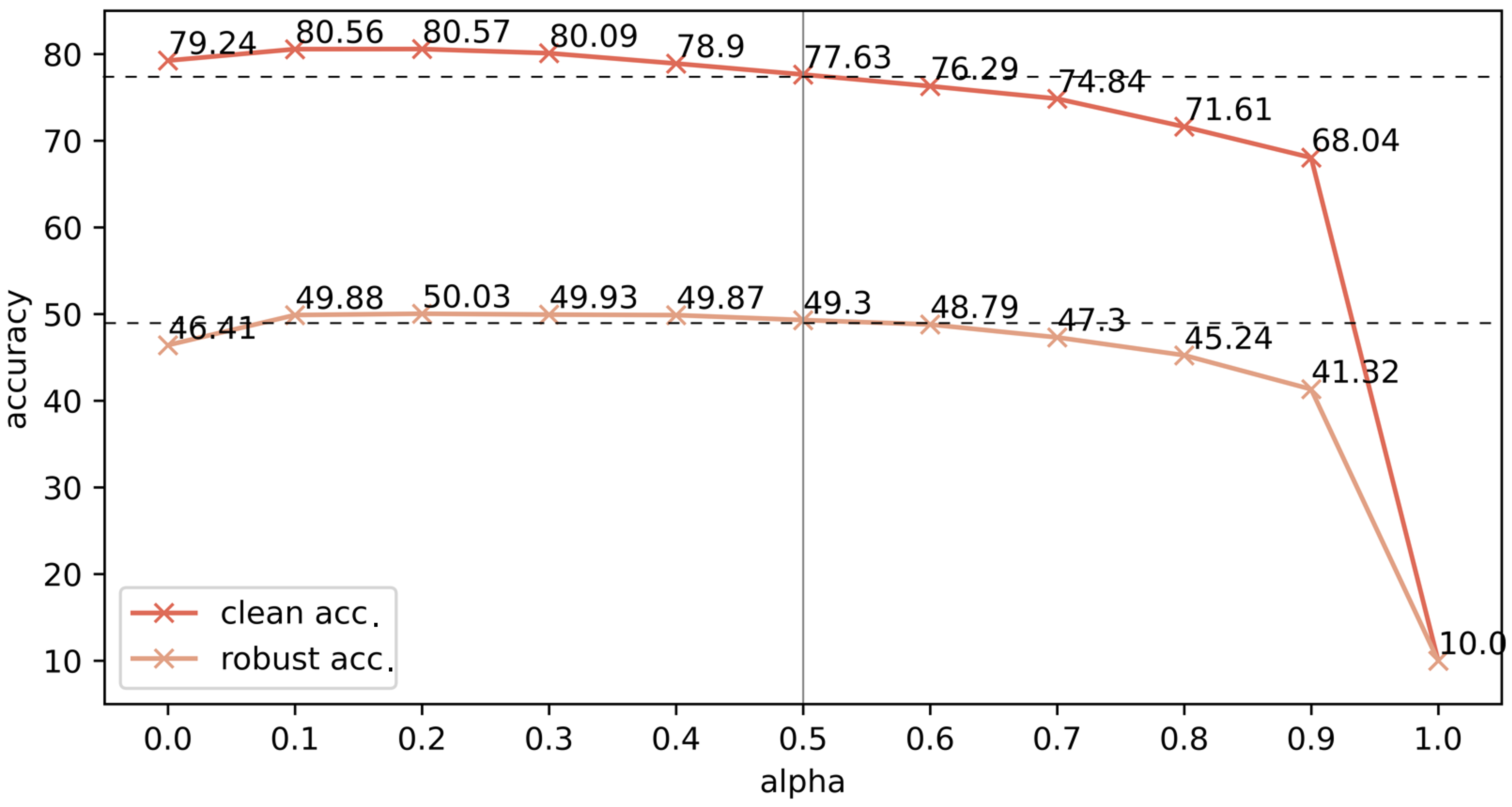}
    \caption{Deep probing for asymmetric similarity function with different $\alpha$.}
    \label{fig:ablation1}
\end{wrapfigure}

We first use fixed $\alpha$ without adaptive annealing to explore the effectiveness of \textit{inferior positives}. Figure~\ref{fig:ablation1}
\setlength{\intextsep}{12pt}
presents the results with different $\alpha$ values when training models for 200 epochs.
Recall that $\alpha$ represents the tendency of the clean sample heading towards the adversarial sample. $\alpha < 0.5$ means clean samples move less toward the adversaries
(vice versa for $\alpha > 0.5$), and $\alpha = 0.5$ degenerates to the original symmetric similarity function form.

Compared with symmetric CoreACL ($\alpha=0.5$), our approach achieves better robustness and accuracy when $\alpha<0.5$ (adversarial examples are treated as \textit{inferior positives}). Intriguingly, when $\alpha=1.0$, the extreme case when only clean samples are attracted by adversaries, we observe the presence of a trivial solution~\cite{chen2021exploring}, that is all images collapse into one point. This validates our observation that adversaries with false identities are indeed pulling their positives towards other instances in the positive contrasts, with the risk of drawing all samples together.
It is also worth noting that when $\alpha < 0.2$, performance begins to drop, showing that a small but non-zero $\alpha$ is the optimal setting empirically.

\subsubsection{Fixed $\alpha$ vs. $\alpha$-Annealing.}
As shown in Table 3, compared to CoreACL, fixed $\alpha$ obtains higher clean accuracy (81.29\% vs. 78.90\%) but with no gain on robust accuracy. Adaptive annealing $\alpha$ achieves both higher robust accuracy (50.24\% vs. 51.27\%) and better clean accuracy (79.46\% vs. 78.90\%).

\setlength{\intextsep}{3pt}  
\begin{wraptable}[12]{r}{5.5cm}
	\centering  
	\fontsize{8.5}{9}\selectfont  
	\begin{threeparttable}
		\caption{Ablation studies, evaluated in SA, RA and time cost. Trained for 400 epochs on 2 Tesla V100 GPUs.}
		\label{tab:headings}  
		\begin{tabular}
        {
        m{1.95cm}
        P{1cm}
        P{1cm}
        P{1.5cm}
        }
			\toprule  
			\makecell[c]{Methods} & SA & RA & Time Cost (s/epoch)\cr  
			\midrule  
			\noalign{\smallskip}
			CoreACL & 78.90 & 50.27 & 96 \\
            \ w/fixed $\alpha$ & 81.29 & 50.24 & 96\\
            \ w/annealing $\alpha$ & 79.46 & 51.37 & 101 \\ \hline
            \ w/$\mathcal{L}^{IP+HN}$ & 81.19 & 51.31 & 101 \\
            AdvCL & 81.35 & 51.00 & 182 \\
			\bottomrule  
		\end{tabular}  
	\end{threeparttable} 
	\label{table:3}
\end{wraptable}
\subsubsection{Comparison with AdvCL.}
Table 3 reports the performance and computation cost comparisons with AdvCL.
CoreACL with $\mathcal{L}^{IP+HN}$ achieves similar performance to AdvCL, which is equivalent to integrate additional components (high frequency view and pseudo-supervision) into CoreACL. The computation time of AdvCL is almost twice than that of $\rm w/\mathcal{L}^{IP+HN}$, which could due to extra computation on contrasting high frequency views and the pseudo-labeled adversarial training.
Our methods only need to compute pair-wise Euclidean distance for $\alpha$-annealing in $\mathcal{L}^{IP}$, and no extra cost introduced in $\mathcal{L}^{HN}$. 

\subsubsection{Effect of Hard Negatives.}
To investigate the effect of hard negatives, we
\setlength{\tabcolsep}{4pt}
\setlength{\intextsep}{6pt}  
\begin{wraptable}[12]{r}{8cm}
\renewcommand\arraystretch{1.5}
	\centering  
	\fontsize{9}{9}\selectfont  
	\scalebox{0.9}{
	\begin{threeparttable}
		\caption{Ablation studies for AdvCL with hard negatives (AdvCL-HN), evaluated under Linear Probing (LP), Adversarial Linear Finetuning (ALF) and Adversarial Full Finetuning (AFF)}  
		\label{tab:performance_comparison}  
		\begin{tabular}{ccccccc} 
			\toprule  
			\multirow{2}{*}{Methods}&  
			\multicolumn{2}{c}{LP}&\multicolumn{2}{c}{ALF}&\multicolumn{2}{c}{AFF}\cr  
			\cmidrule(lr){2-3} \cmidrule(lr){4-5}  \cmidrule(lr){6-7}  
			&SA&RA&SA&RA&SA&RA\cr  
			\midrule  
             
			{AdvCL-HN}&{81.34}&{\bf 52.96}&{78.69}&{\bf 53.20}&83.44&{\bf 54.07}\cr  
			{w/o debias}&{\bf 81.52}&51.61&{\bf 78.89}&52.34&{\bf 83.73}&{54.01}\cr 
		{w/o reweight}&76.93&50.01&73.49&49.86&81.74&52.60\cr 
			\bottomrule  
		\end{tabular}  
	\end{threeparttable}}  
\end{wraptable}
evaluate each component (negatives debiasing~\cite{chuang2020debiased}, reweighting~\cite{robinson2020contrastive}) as shown in Table 4. With negatives-debiasing removed, we observe decrease in robust accuracy, with slightly increased standard accuracy. We hypothesize that without debiasing, semantically similar adversarial representations that should be mapped closely are pushed away instead. 
In addition, the removal of negatives reweighting results in a sharp performance drop, showing that viewing adversarial views as \textit{hard negatives} with higher weights plays a key role in discriminating adversarial samples.

\begin{figure}[t]
    \centering
    \includegraphics[scale=0.42]{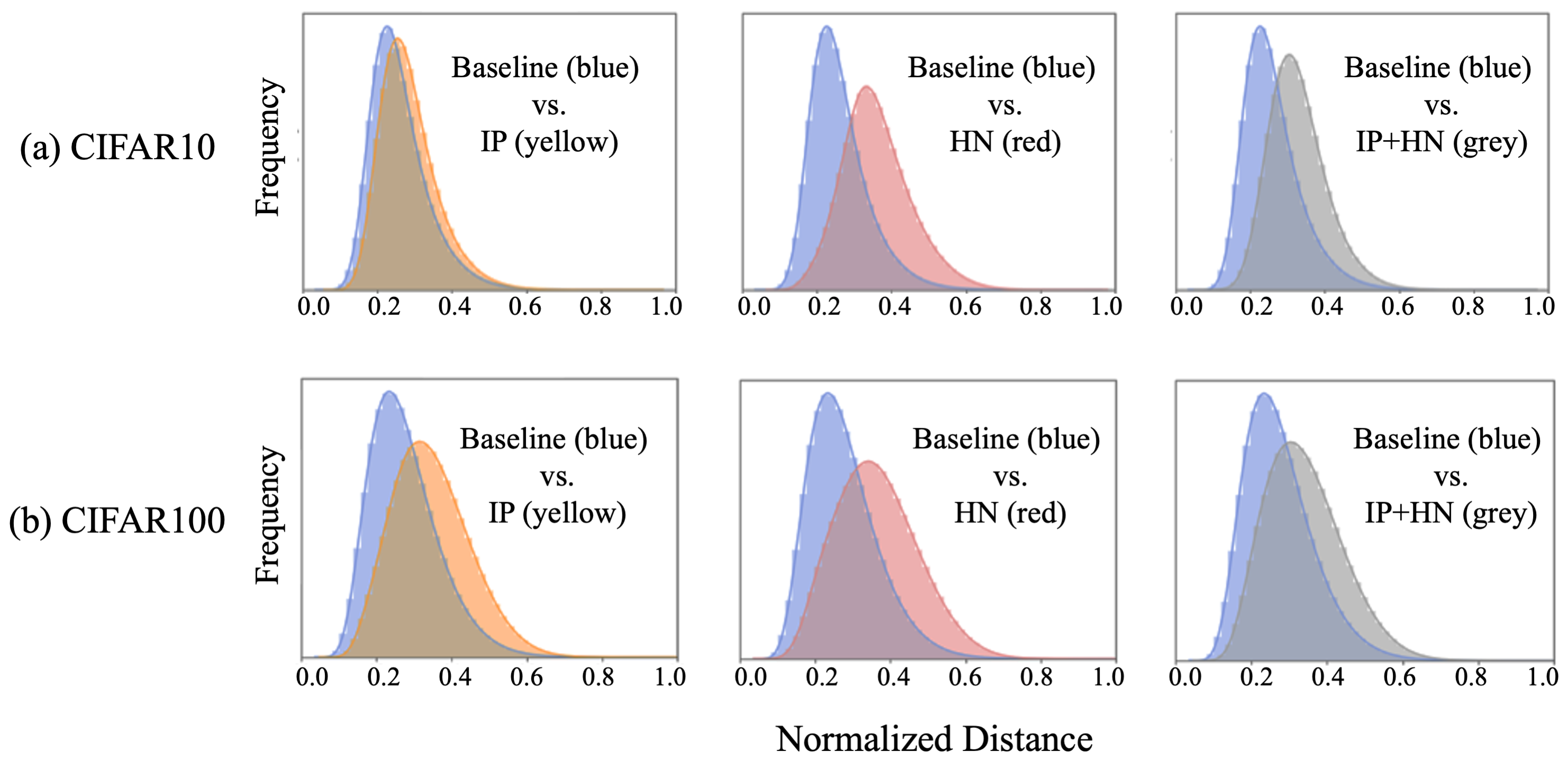}
    \caption{Histograms of Euclidean distance (normalized) distribution of all negative pairs learned by different objectives in (a) CIFAR10 (first row) and (b) CIFAR100 (second row). Baseline is AdvCL~\cite{fan2021does}; IP: baseline with Inferior Positives; HN: baseline with Hard Negatives. On each dataset, our methods are better at differentiating different instances (with larger distance between negative pairs)
    }
    \label{fig:ablation2}
\end{figure}

\begin{figure}[b]
    \centering
    \includegraphics[scale=0.39]{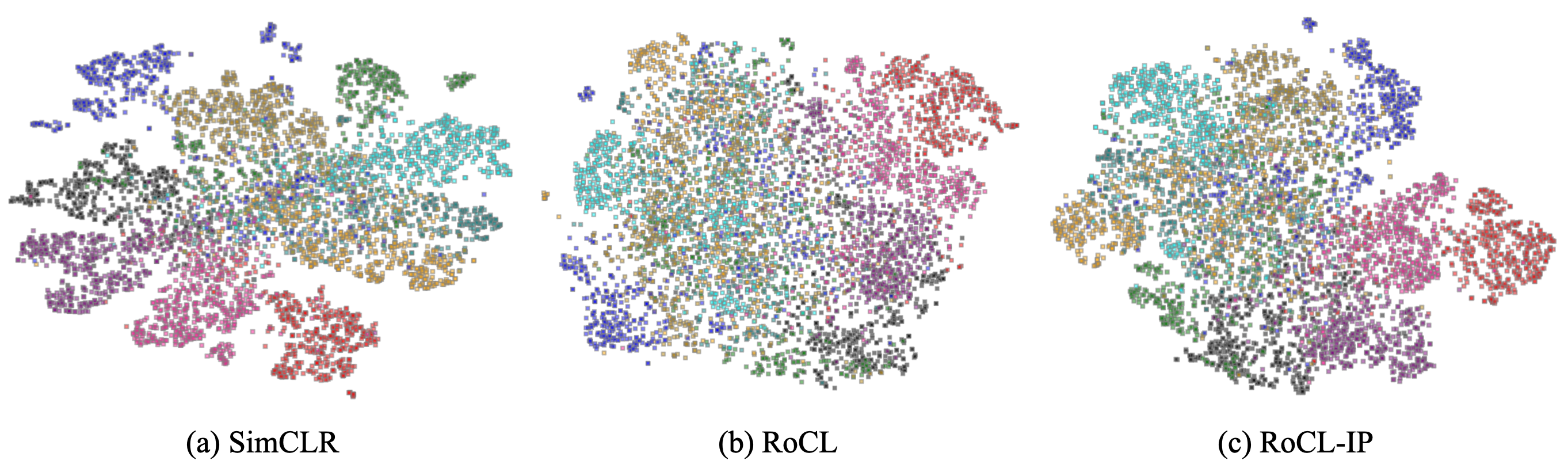}
    \caption{t-SNE visualizations in a global view on CIFAR-10 validation set. The embeddings are learned by different self-supervised pre-training methods (SimCLR(a), RoCL(b) and RoCL-IP(c)) (\textit{colored figure})
    }
    \label{fig:global tsne}
\end{figure}

\subsection{Qualitative Analysis}

Figure \ref{fig:ablation2} shows the distribution of normalized Euclidean distance over all negative pairs. We take AdvCL~\cite{fan2021does} as the baseline and compare it with its enhanced versions with our methods.
Generally, our methods can shift the original distribution curve right (larger distance), meaning that treating adversaries as inferior positives or hard negatives encourages the model to separate negative pairs further apart and induce better instance discrimination. This suggests that our proposed methods effectively mitigate the negative impacts of \textit{identity confusion}.

Figure \ref{fig:global tsne} provides 2-D visualization (t-SNE~\cite{van2008visualizing} on CIFAR-10) for the embeddings learnt by SimCLR~\cite{pmlr-v119-chen20j}, RoCL~\cite{kim2020adversarial} and RoCL enhanced by $\mathcal{L}^{IP}$ (RoCL-IP). Each class is represented in one color.
Compared to SimCLR, RoCL representations are corrupted by adversaries and exhibit poor class discrimination. RoCL-IP yields better class separation compared with RoCL.
This shows that asymmetric similarity consideration eases instance-level identity confusion.


\section{Related Work}

\paragraph{Contrastive Learning}

CL has been widely applied to learn generalizable features from unlabeled data~\cite{pmlr-v119-chen20j,he2020momentum,tian2020contrastive,grill2020bootstrap,chen2020improved,caron2020unsupervised,bachman2019learning,oord2018representation,chen2020big,chen2021large,khosla2020supervised}. The basic idea is instance discrimination~\cite{wu2018unsupervised}.
Representative works include CMC~\cite{tian2020contrastive}, SimCLR\cite{pmlr-v119-chen20j}
, MoCo\cite{he2020momentum}, SwAV\cite{caron2020unsupervised}, BYOL\cite{grill2020bootstrap}. 
There is also a stream of work focusing on refined sampling on different views for improved performance ~\cite{tian2020contrastive,kalantidis2020hard,chuang2020debiased,robinson2020contrastive,tao2021clustering}.
For example, DCL\cite{chuang2020debiased} proposed to \textit{debias} the assumption that all negative pairs are true negatives. HCL\cite{robinson2020contrastive} extended DCL and proposed to mine hard negatives for contrastive learning, whose embeddings are uneasy to discriminate. 

\paragraph{Adversarial Training}

Adversarial training (AT) stems from \cite{goodfellow2014explaining} and adopts a min-max training regime that optimizes the objective over adversaries generated by maximizing the loss~\cite{madry2018towards,zhang2019theoretically,shafahi2019adversarial,zhang2019you,wong2020fast,zhu2019freelb,gan2020large,pang2019rethinking}. Some recent work introduced unlabeled data into AT~\cite{hendrycks2019using,chen2020adversarial,carmon2019unlabeled,alayrac2019labels,kim2020adversarial}. By leveraging a large amount of unlabeled data, \cite{carmon2019unlabeled,alayrac2019labels} performed semi-supervised self-training to first generate pseudo-supervisions, then conducted conventional supervised AT. Our work explores how to learn robust models without any class labels.

\paragraph{Adversarial Contrastive Learning}

Some recent studies applied CL on adversarial training~\cite{kim2020adversarial,jiang2020robust,fan2021does,gowal2020self}, by considering adversaries as positive views for contrasting, such that the learned encoder renders robust data representations. RoCL~\cite{kim2020adversarial} was the first to successfully show robust models can be learned in an unsupervised manner.
AdvCL~\cite{fan2021does} proposed to empower CL with pseudo-supervision stimulus.
Same as CL, these Adversarial CL methods perform symmetric contrast for all pairs, which could potentially induces conflicts in CL and AT training objectives. We are the first to investigate the asymmetric properties of Adversarial CL, by treating adversaries discriminatingly.

\section{Conclusions}

In this work, we study enhancing model robustness using unlabeled data and investigate the \textit{identity confusion} issue in Adversarial CL, \textit{i.e.}, adversaries with different identities attract their anchors together, contradicting to the objective of CL. 
We present a generic asymmetric objective \textit{A-InfoNCE}, and treat adversaries discriminatingly as \textit{inferior positives} or \textit{hard negatives}, which can overcome the identify confusion challenge.
Comprehensive experiments with quantitative and qualitative analysis show that our methods can enhance existing Adversarial CL methods effectively.
Further, it lies in our future work to extend the proposed asymmetric form to other CL settings to take into consideration the asymmetric characteristics between different views.

\section*{Acknowledgement}

This work was supported in part by the National Key R\&D Program of China under Grant 2021ZD0112100, partly by Baidu Inc. through Apollo-AIR Joint Research Center. We would also like to thank the anonymous reviewers for their insightful comments.

\clearpage
%
%
\bibliographystyle{splncs04}
\bibliography{egbib}

\begin{thebibliography}{10}
\providecommand{\url}[1]{\texttt{#1}}
\providecommand{\urlprefix}{URL }
\providecommand{\doi}[1]{https://doi.org/#1}

\bibitem{alayrac2019labels}
Alayrac, J.B., Uesato, J., Huang, P.S., Fawzi, A., Stanforth, R., Kohli, P.:
  Are labels required for improving adversarial robustness? Advances in Neural
  Information Processing Systems  \textbf{32} (2019)

\bibitem{athalye2018obfuscated}
Athalye, A., Carlini, N., Wagner, D.: Obfuscated gradients give a false sense
  of security: Circumventing defenses to adversarial examples. In:
  International conference on machine learning. pp. 274--283. PMLR (2018)

\bibitem{bachman2019learning}
Bachman, P., Hjelm, R.D., Buchwalter, W.: Learning representations by
  maximizing mutual information across views. Advances in neural information
  processing systems  \textbf{32} (2019)

\bibitem{carlini2017towards}
Carlini, N., Wagner, D.: Towards evaluating the robustness of neural networks.
  In: 2017 ieee symposium on security and privacy (sp). pp. 39--57. IEEE (2017)

\bibitem{carmon2019unlabeled}
Carmon, Y., Raghunathan, A., Schmidt, L., Duchi, J.C., Liang, P.S.: Unlabeled
  data improves adversarial robustness. Advances in Neural Information
  Processing Systems  \textbf{32} (2019)

\bibitem{caron2020unsupervised}
Caron, M., Misra, I., Mairal, J., Goyal, P., Bojanowski, P., Joulin, A.:
  Unsupervised learning of visual features by contrasting cluster assignments.
  Advances in Neural Information Processing Systems  \textbf{33},  9912--9924
  (2020)

\bibitem{chen2021large}
Chen, S., Niu, G., Gong, C., Li, J., Yang, J., Sugiyama, M.: Large-margin
  contrastive learning with distance polarization regularizer. In:
  International Conference on Machine Learning. pp. 1673--1683. PMLR (2021)

\bibitem{chen2020adversarial}
Chen, T., Liu, S., Chang, S., Cheng, Y., Amini, L., Wang, Z.: Adversarial
  robustness: From self-supervised pre-training to fine-tuning. In: Proceedings
  of the IEEE/CVF Conference on Computer Vision and Pattern Recognition. pp.
  699--708 (2020)

\bibitem{pmlr-v119-chen20j}
Chen, T., Kornblith, S., Norouzi, M., Hinton, G.: A simple framework for
  contrastive learning of visual representations. In: III, H.D., Singh, A.
  (eds.) Proceedings of the 37th International Conference on Machine Learning.
  Proceedings of Machine Learning Research, vol.~119, pp. 1597--1607. PMLR
  (13--18 Jul 2020), \url{https://proceedings.mlr.press/v119/chen20j.html}

\bibitem{chen2020big}
Chen, T., Kornblith, S., Swersky, K., Norouzi, M., Hinton, G.E.: Big
  self-supervised models are strong semi-supervised learners. Advances in
  neural information processing systems  \textbf{33},  22243--22255 (2020)

\bibitem{chen2020improved}
Chen, X., Fan, H., Girshick, R., He, K.: Improved baselines with momentum
  contrastive learning. arXiv preprint arXiv:2003.04297  (2020)

\bibitem{chen2021exploring}
Chen, X., He, K.: Exploring simple siamese representation learning. In:
  Proceedings of the IEEE/CVF Conference on Computer Vision and Pattern
  Recognition. pp. 15750--15758 (2021)

\bibitem{chuang2020debiased}
Chuang, C.Y., Robinson, J., Lin, Y.C., Torralba, A., Jegelka, S.: Debiased
  contrastive learning. Advances in neural information processing systems
  \textbf{33},  8765--8775 (2020)

\bibitem{croce2020reliable}
Croce, F., Hein, M.: Reliable evaluation of adversarial robustness with an
  ensemble of diverse parameter-free attacks. In: International conference on
  machine learning. pp. 2206--2216. PMLR (2020)

\bibitem{dong2018boosting}
Dong, Y., Liao, F., Pang, T., Su, H., Zhu, J., Hu, X., Li, J.: Boosting
  adversarial attacks with momentum. In: Proceedings of the IEEE conference on
  computer vision and pattern recognition. pp. 9185--9193 (2018)

\bibitem{du2014analysis}
Du~Plessis, M.C., Niu, G., Sugiyama, M.: Analysis of learning from positive and
  unlabeled data. Advances in neural information processing systems
  \textbf{27} (2014)

\bibitem{elkan2008learning}
Elkan, C., Noto, K.: Learning classifiers from only positive and unlabeled
  data. In: Proceedings of the 14th ACM SIGKDD international conference on
  Knowledge discovery and data mining. pp. 213--220 (2008)

\bibitem{fan2021does}
Fan, L., Liu, S., Chen, P.Y., Zhang, G., Gan, C.: When does contrastive
  learning preserve adversarial robustness from pretraining to finetuning?
  Advances in Neural Information Processing Systems  \textbf{34} (2021)

\bibitem{gan2020large}
Gan, Z., Chen, Y.C., Li, L., Zhu, C., Cheng, Y., Liu, J.: Large-scale
  adversarial training for vision-and-language representation learning.
  Advances in Neural Information Processing Systems  \textbf{33},  6616--6628
  (2020)

\bibitem{goodfellow2014explaining}
Goodfellow, I.J., Shlens, J., Szegedy, C.: Explaining and harnessing
  adversarial examples. arXiv preprint arXiv:1412.6572  (2014)

\bibitem{gowal2020self}
Gowal, S., Huang, P.S., van~den Oord, A., Mann, T., Kohli, P.: Self-supervised
  adversarial robustness for the low-label, high-data regime. In: International
  Conference on Learning Representations (2020)

\bibitem{grill2020bootstrap}
Grill, J.B., Strub, F., Altch{\'e}, F., Tallec, C., Richemond, P., Buchatskaya,
  E., Doersch, C., Avila~Pires, B., Guo, Z., Gheshlaghi~Azar, M., et~al.:
  Bootstrap your own latent-a new approach to self-supervised learning.
  Advances in Neural Information Processing Systems  \textbf{33},  21271--21284
  (2020)

\bibitem{hadsell2006dimensionality}
Hadsell, R., Chopra, S., LeCun, Y.: Dimensionality reduction by learning an
  invariant mapping. In: 2006 IEEE Computer Society Conference on Computer
  Vision and Pattern Recognition (CVPR'06). vol.~2, pp. 1735--1742. IEEE (2006)

\bibitem{he2020momentum}
He, K., Fan, H., Wu, Y., Xie, S., Girshick, R.: Momentum contrast for
  unsupervised visual representation learning. In: Proceedings of the IEEE/CVF
  Conference on Computer Vision and Pattern Recognition. pp. 9729--9738 (2020)

\bibitem{he2016deep}
He, K., Zhang, X., Ren, S., Sun, J.: Deep residual learning for image
  recognition. In: Proceedings of the IEEE conference on computer vision and
  pattern recognition. pp. 770--778 (2016)

\bibitem{hendrycks2019using}
Hendrycks, D., Mazeika, M., Kadavath, S., Song, D.: Using self-supervised
  learning can improve model robustness and uncertainty. Advances in Neural
  Information Processing Systems  \textbf{32},  15663--15674 (2019)

\bibitem{jiang2020robust}
Jiang, Z., Chen, T., Chen, T., Wang, Z.: Robust pre-training by adversarial
  contrastive learning. In: NeurIPS (2020)

\bibitem{kalantidis2020hard}
Kalantidis, Y., Sariyildiz, M.B., Pion, N., Weinzaepfel, P., Larlus, D.: Hard
  negative mixing for contrastive learning. Advances in Neural Information
  Processing Systems  \textbf{33},  21798--21809 (2020)

\bibitem{kannan2018adversarial}
Kannan, H., Kurakin, A., Goodfellow, I.: Adversarial logit pairing. arXiv
  preprint arXiv:1803.06373  (2018)

\bibitem{kantipudi2020color}
Kantipudi, J., Dubey, S.R., Chakraborty, S.: Color channel perturbation attacks
  for fooling convolutional neural networks and a defense against such attacks.
  IEEE Transactions on Artificial Intelligence  \textbf{1}(2),  181--191 (2020)

\bibitem{khosla2020supervised}
Khosla, P., Teterwak, P., Wang, C., Sarna, A., Tian, Y., Isola, P., Maschinot,
  A., Liu, C., Krishnan, D.: Supervised contrastive learning. Advances in
  Neural Information Processing Systems  \textbf{33},  18661--18673 (2020)

\bibitem{kim2020adversarial}
Kim, M., Tack, J., Hwang, S.J.: Adversarial self-supervised contrastive
  learning. Advances in Neural Information Processing Systems  \textbf{33}
  (2020)

\bibitem{li2020prototypical}
Li, J., Zhou, P., Xiong, C., Hoi, S.C.: Prototypical contrastive learning of
  unsupervised representations. arXiv preprint arXiv:2005.04966  (2020)

\bibitem{van2008visualizing}
Van~der Maaten, L., Hinton, G.: Visualizing data using t-sne. Journal of
  machine learning research  \textbf{9}(11) (2008)

\bibitem{madry2018towards}
Madry, A., Makelov, A., Schmidt, L., Tsipras, D., Vladu, A.: Towards deep
  learning models resistant to adversarial attacks. In: International
  Conference on Learning Representations (2018)

\bibitem{oord2018representation}
Oord, A.v.d., Li, Y., Vinyals, O.: Representation learning with contrastive
  predictive coding. arXiv preprint arXiv:1807.03748  (2018)

\bibitem{pang2019rethinking}
Pang, T., Xu, K., Dong, Y., Du, C., Chen, N., Zhu, J.: Rethinking softmax
  cross-entropy loss for adversarial robustness. arXiv preprint
  arXiv:1905.10626  (2019)

\bibitem{rahamim2022robustness}
Rahamim, A., Naeh, I.: Robustness through cognitive dissociation mitigation in
  contrastive adversarial training. arXiv preprint arXiv:2203.08959  (2022)

\bibitem{robinson2020contrastive}
Robinson, J.D., Chuang, C.Y., Sra, S., Jegelka, S.: Contrastive learning with
  hard negative samples. In: International Conference on Learning
  Representations (2020)

\bibitem{shafahi2019adversarial}
Shafahi, A., Najibi, M., Ghiasi, M.A., Xu, Z., Dickerson, J., Studer, C.,
  Davis, L.S., Taylor, G., Goldstein, T.: Adversarial training for free!
  Advances in Neural Information Processing Systems  \textbf{32} (2019)

\bibitem{su2019one}
Su, J., Vargas, D.V., Sakurai, K.: One pixel attack for fooling deep neural
  networks. IEEE Transactions on Evolutionary Computation  \textbf{23}(5),
  828--841 (2019)

\bibitem{szegedy2014intriguing}
Szegedy, C., Zaremba, W., Sutskever, I., Bruna, J., Erhan, D., Goodfellow, I.,
  Fergus, R.: Intriguing properties of neural networks. In: 2nd International
  Conference on Learning Representations, ICLR 2014 (2014)

\bibitem{tao2021clustering}
Tao, Y., Takagi, K., Nakata, K.: Clustering-friendly representation learning
  via instance discrimination and feature decorrelation. arXiv preprint
  arXiv:2106.00131  (2021)

\bibitem{tian2020contrastive}
Tian, Y., Krishnan, D., Isola, P.: Contrastive multiview coding. In: European
  conference on computer vision. pp. 776--794. Springer (2020)

\bibitem{tian2020makes}
Tian, Y., Sun, C., Poole, B., Krishnan, D., Schmid, C., Isola, P.: What makes
  for good views for contrastive learning? Advances in Neural Information
  Processing Systems  \textbf{33},  6827--6839 (2020)

\bibitem{wahed2022adversarial}
Wahed, M., Tabassum, A., Lourentzou, I.: Adversarial contrastive learning by
  permuting cluster assignments. arXiv preprint arXiv:2204.10314  (2022)

\bibitem{wong2020fast}
Wong, E., Rice, L., Kolter, J.Z.: Fast is better than free: Revisiting
  adversarial training. arXiv preprint arXiv:2001.03994  (2020)

\bibitem{wu2018unsupervised}
Wu, Z., Xiong, Y., Yu, S.X., Lin, D.: Unsupervised feature learning via
  non-parametric instance discrimination. In: Proceedings of the IEEE
  conference on computer vision and pattern recognition. pp. 3733--3742 (2018)

\bibitem{xiao2018spatially}
Xiao, C., Zhu, J.Y., Li, B., He, W., Liu, M., Song, D.: Spatially transformed
  adversarial examples. arXiv preprint arXiv:1801.02612  (2018)

\bibitem{zhang2019you}
Zhang, D., Zhang, T., Lu, Y., Zhu, Z., Dong, B.: You only propagate once:
  Accelerating adversarial training via maximal principle. Advances in Neural
  Information Processing Systems  \textbf{32} (2019)

\bibitem{zhang2019theoretically}
Zhang, H., Yu, Y., Jiao, J., Xing, E., El~Ghaoui, L., Jordan, M.: Theoretically
  principled trade-off between robustness and accuracy. In: International
  conference on machine learning. pp. 7472--7482. PMLR (2019)

\bibitem{zhu2019freelb}
Zhu, C., Cheng, Y., Gan, Z., Sun, S., Goldstein, T., Liu, J.: Freelb: Enhanced
  adversarial training for natural language understanding. arXiv preprint
  arXiv:1909.11764  (2019)

\end{thebibliography}

\clearpage

\section*{Appendix}

\renewcommand{\thesubsection}{\Alph{subsection}}

\subsection{Results under more attacks}

In order to verify the effectiveness of the proposed method, in this section, we further evaluate the robustness of our method under a broader range of powerful attacks: $1)$ AutoAttack~\cite{croce2020reliable} (an ensemble of four strong diverse attacks, which is widely considered as the strongest attack for robustness evaluation), $2)$ CW attack~\cite{carlini2017towards} (CW-200), $3)$ PGD attack with restart~\cite{madry2018towards} (PGD-200), $4)$ One-pixel attack~\cite{su2019one}, $5)$ Spatial Transformation attack~\cite{xiao2018spatially}, as well as $6)$ Color Channel attack~\cite{kantipudi2020color}. PGD-200 and CW-200 both restart 5 times with 40 optimization steps each restart.

In Table~\ref{tab:other_attacks-1}, we report the robust accuracy under these attacks with AdvCL serving as baseline on CIFAR100.
The results show that our methods can improve robustness under all different attacks across almost all settings, \textit{e.g.}, 21.43\% vs. 19.57\% under AutoAttack and 29.56\% vs. 27.13\% under PGD-200 attack, with loss function $\mathcal{L}^{IP+HN}$ (Equation~\ref{eq:ip+hn}), under Linear Probing.

\begin{table}[h]
\fontsize{8}{9}\selectfont 
\renewcommand\arraystretch{1.5}
\centering
\caption{Robustness evaluation under diverse attacks on CIFAR100 with AdvCL as baseline.}
\vspace{5pt}
\label{tab:other_attacks-1}
\scalebox{0.88}{
\begin{tabular}{clcccccc}
\toprule
\multicolumn{2}{c}{\begin{tabular}[c]{@{}c@{}} Training Methods\end{tabular}} & PGD-200 & CW-200 & AA & One-pix. & Spatial-Tr. & Color-Ch. \\ \hline\hline
\multirow{4}{*}{Linear Probing} & AdvCL & 27.13 & 21.85 & 19.57 & 72.10 & 47.94 & 25.62 \\
 & \ \ w/ $\mathcal{L}^{IP}$ & 27.87 & 22.10 & 19.80 & 69.60 & 49.31 & 25.88 \\
 & \ \ w/ $\mathcal{L}^{HN}$ & 29.43 & 23.10 & 21.23 & \textbf{73.20} & 51.57 & 28.01 \\
 & \ \ w/ $\mathcal{L}^{IP+HN}$ & \textbf{29.56} & \textbf{23.60} & \textbf{21.43} & 73.00 & \textbf{52.62} & \textbf{28.94} \\ \hline
\multirow{4}{*}{\begin{tabular}[c]{@{}c@{}}Adversarial Linear\\ Finetuning\end{tabular}} & AdvCL & 27.29 & 22.01 & 20.09 & \textbf{72.80} & 47.31 & 24.98 \\
 & \ \ w/ $\mathcal{L}^{IP}$ & 27.84 & 22.37 & 20.06 & 71.60 & 46.22 & 24.23 \\
 & \ \ w/ $\mathcal{L}^{HN}$ & \textbf{29.79} & \textbf{23.79} & 21.52 & 70.80 & \textbf{51.04} & \textbf{27.84} \\
 & \ \ w/ $\mathcal{L}^{IP+HN}$ & 29.58 & 23.64 & \textbf{21.66} & 71.70 & 49.87 & 27.14 \\ \hline
\multirow{4}{*}{\begin{tabular}[c]{@{}c@{}}Adversarial Full\\ Finetuning\end{tabular}} & AdvCL & 29.48 & 25.73 & 24.46 & \textbf{72.20} & 57.86 & 25.12 \\
 & \ \ w/ $\mathcal{L}^{IP}$ & 30.10 & 26.05 & 24.73 & 71.00 & 58.95 & 25.55 \\
 & \ \ w/ $\mathcal{L}^{HN}$ & \textbf{30.46} & \textbf{26.60} & \textbf{25.22} & 69.30 & 59.04 & \textbf{26.02} \\
 & \ \ w/ $\mathcal{L}^{IP+HN}$ & \textbf{30.46} & 26.54 & 25.06 & 69.00 & \textbf{59.33} & 25.70 \\ \toprule
\end{tabular}}
\end{table}

Table~\ref{tab:other_attacks-2} provides results on CIFAR10 under canonical optimization-based attack methods: PGD-200, CW-200 and AutoAttack. Our methods also yield robustness gain in almost all settings. 

\begin{table}[h]
\fontsize{8}{9}\selectfont 
\renewcommand\arraystretch{1.5}
\centering
\vspace{10pt}
\caption{Robustness evaluation under optimization-based attacks on CIFAR10, with AdvCL as baseline.}
\vspace{5pt}
\label{tab:other_attacks-2}
\begin{tabular}{clccc}
\toprule
\multicolumn{2}{c}{\begin{tabular}[c]{@{}c@{}}Training Methods\end{tabular}} & PGD-200 & CW-200 & AutoAttack \\ \hline\hline
\multirow{4}{*}{Linear Probing} & AdvCL & 51.05 & 45.65 & 43.48 \\
 & \ \ w/ $\mathcal{L}^{IP}$ & 51.99 & 46.02 & 43.57 \\
 & \ \ w/ $\mathcal{L}^{HN}$ & \textbf{52.36} & \textbf{46.09} & \textbf{43.68} \\
 & \ \ w/ $\mathcal{L}^{IP+HN}$ & 52.01 & 45.35 & 42.92 \\ \hline
\multirow{4}{*}{\begin{tabular}[c]{@{}c@{}}Adversarial Linear\\ Finetuning\end{tabular}} & AdvCL & 52.30 & 46.04 & 43.93 \\
 & \ \ w/ $\mathcal{L}^{IP}$ & 52.77 & 46.60 & \textbf{44.22} \\
 & \ \ w/ $\mathcal{L}^{HN}$ & \textbf{53.22} & \textbf{46.44} & 44.15 \\
 & \ \ w/ $\mathcal{L}^{IP+HN}$ & 52.77 & 45.55 & 43.01 \\ \hline
\multirow{4}{*}{\begin{tabular}[c]{@{}c@{}}Adversarial Full\\ Finetuning\end{tabular}} & AdvCL & 52.90 & 50.92 & 49.58 \\
 & \ \ w/ $\mathcal{L}^{IP}$ & \textbf{53.61} & 51.25 & 49.90 \\
 & \ \ w/ $\mathcal{L}^{HN}$ & 53.25 & 51.11 & 49.93 \\
 & \ \ w/ $\mathcal{L}^{IP+HN}$ & 53.51 & \textbf{51.46} & \textbf{50.28} \\ \toprule
\end{tabular}
\end{table}
\vspace{5pt}

Besides, we also report results compared with RoCL under PGD-200, CW-200 and AutoAttack in Table~\ref{tab:rocl-other_attacks}, which further validate the effectiveness of the proposed methods. For instance, 25.09\% vs. 23.51\% under CW-200 attack, Adversarial Full Finetuning scheme, on CIFAR100.

\begin{table}[h]
\fontsize{8}{9}\selectfont 
\renewcommand\arraystretch{1.7}
\centering
\vspace{10pt}
\caption{Robustness evaluation under optimization-based attacks, with RoCL as baseline, on CIFAR-10 and CIFAR-100.}
\vspace{5pt}
\label{tab:rocl-other_attacks}
\begin{tabular}{cclccc}
\toprule
\multicolumn{1}{l}{Dataset} & \multicolumn{2}{c}{\begin{tabular}[c]{@{}c@{}}Training Methods\end{tabular}} & PGD-200 & CW-200 & AutoAttack \\ \hline\hline
\multirow{6}{*}{CIFAR10} & \multirow{2}{*}{Linear Probing} & RoCL & 32.47 & 33.33 & 24.11 \\
 &  & \ \ w/ $\mathcal{L}^{IP+HN}$ & \textbf{34.13} & \textbf{34.59} & \textbf{24.58} \\ \cline{3-6} 
 & \multirow{2}{*}{\begin{tabular}[c]{@{}c@{}}Adversarial Linear\\ Finetuning\end{tabular}} & RoCL & 42.58 & 40.21 & \textbf{31.81} \\
 &  & \ \ w/ $\mathcal{L}^{IP+HN}$ & \textbf{43.54} & \textbf{41.26} & 30.37 \\ \cline{3-6} 
 & \multirow{2}{*}{\begin{tabular}[c]{@{}c@{}}Adversarial Full\\ Finetuning\end{tabular}} & RoCL & 50.33 & 47.57 & 46.69 \\
 &  & \ \ w/ $\mathcal{L}^{IP+HN}$ & \textbf{51.47} & \textbf{48.26} & \textbf{47.05} \\ \hline
\multirow{6}{*}{CIFAR100} & \multirow{2}{*}{Linear Probing} & RoCL & 14.93 & 14.75 & 7.58 \\
 &  & \ \ w/ $\mathcal{L}^{IP+HN}$ & \textbf{17.95} & \textbf{16.57} & \textbf{8.58} \\ \cline{3-6} 
 & \multirow{2}{*}{\begin{tabular}[c]{@{}c@{}}Adversarial Linear\\ Finetuning\end{tabular}} & RoCL & 22.59 & 18.99 & \textbf{11.93} \\
 &  & \ \ w/ $\mathcal{L}^{IP+HN}$ & \textbf{24.46} & \textbf{20.69} & 11.69 \\ \cline{3-6} 
 & \multirow{2}{*}{\begin{tabular}[c]{@{}c@{}}Adversarial Full\\ Finetuning\end{tabular}} & RoCL & 27.95 & 23.51 & 22.70 \\
 &  & \ \ w/ $\mathcal{L}^{IP+HN}$ & \textbf{29.37} & \textbf{25.09} & \textbf{24.01} \\ \toprule
\end{tabular}
\end{table}
\vspace{5pt}
\end{document}